\documentclass[10pt,twocolumn,letterpaper]{article}

\usepackage{cvpr}
\usepackage{times}
\usepackage{epsfig}
\usepackage{graphicx}
\usepackage{amsmath}
\usepackage{amssymb}
\usepackage{cite} 
\usepackage{kotex}
\usepackage{dsfont}
\usepackage{adjustbox}
\usepackage{balance}
\usepackage[linesnumbered,ruled,vlined]{algorithm2e}
\usepackage{ctable}
\usepackage{multirow}
\usepackage{makecell}
\usepackage[symbol]{footmisc}
\usepackage{float}

\usepackage[T1]{fontenc}
\usepackage{etoolbox}
 
\makeatletter
\patchcmd{\maketitle}{}{}  

\makeatother
\makeatletter
\def\@fnsymbol#1{\ensuremath{\ifcase#1\or \dagger\or \ddagger\or
    \mathsection\or \mathparagraph\or \|\or **\or \dagger\dagger
    \or \ddagger\ddagger \else\@ctrerr\fi}}
\makeatother

\usepackage[pagebackref=true,breaklinks=true,letterpaper=true,colorlinks,bookmarks=false]{hyperref}

\cvprfinalcopy 

\def\cvprPaperID{****} 

\ifcvprfinal\fi
\begin{document}

\title{FickleNet: Weakly and Semi-supervised Semantic Image Segmentation\\using Stochastic Inference}
\author{Jungbeom Lee~~~~~~~ Eunji Kim~~~~~~~ Sungmin Lee ~~~~~~~ Jangho Lee ~~~~~~~ Sungroh Yoon\thanks{Correspondence to: Sungroh Yoon <sryoon@snu.ac.kr>.}\\
Department of Electrical and Computer Engineering, Seoul National University, Seoul, South Korea\\
{\tt\small \{jbeom.lee93, kce407, simonlee0810, ubuntu, sryoon\}@snu.ac.kr}}

\maketitle

\begin{abstract}

The main obstacle to weakly supervised semantic image segmentation is the difficulty of obtaining pixel-level information from coarse image-level annotations. Most methods based on image-level annotations use localization maps obtained from the classifier, but these only focus on the small discriminative parts of objects and do not capture precise boundaries.
FickleNet explores diverse combinations of locations on feature maps created by generic deep neural networks. It selects hidden units randomly and then uses them to obtain activation scores for image classification.
FickleNet implicitly learns the coherence of each location in the feature maps, resulting in a localization map which identifies both discriminative and other parts of objects.
The ensemble effects are obtained from a single network by selecting random hidden unit pairs, which means that a variety of localization maps are generated from a single image.
Our approach does not require any additional training steps and only adds a simple layer to a standard convolutional neural network; nevertheless it outperforms recent comparable techniques on the Pascal VOC 2012 benchmark in both weakly and semi-supervised settings.
\end{abstract}
\begin{figure}[t]
  \centering
  \includegraphics[width=1.0\linewidth]{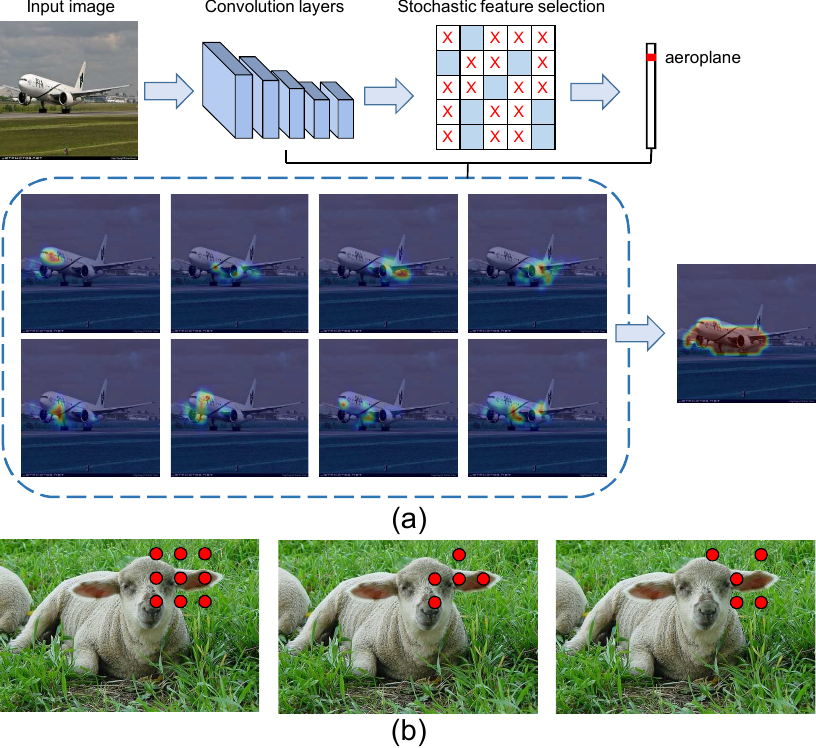}
  \caption{\label{fig_1} (a) FickleNet allows a single network to generate multiple localization maps from a single image. (b) Conceptual description of hidden unit selection. Selecting all hidden units (deterministic, \textit{left}) produces smoothing effects as background and foreground are activated together. Randomly selected hidden units (stochastic, \textit{center} and \textit{right}) can provide more flexible combinations which can correspond more clearly to parts of objects or the background.}
  \vspace*{-10pt}
  \end{figure}
\section{Introduction}
Semantic segmentation is one of the most important and interesting tasks in computer vision, and the development of deep learning has produced tremendous progress in a fully supervised setting~\cite{zhao2017pyramid,chen2014semantic}. 
However, to use semantic image segmentation in real life requires a large variety of object classes and a great deal of labeled data for each class.
Labeling pixel-level annotations of each object class is laborious, and hampers the expansion of object classes. 
This problem can be addressed by weakly supervised methods that use annotations, which are less definite than those at the pixel level and much easier to obtain.
However, current weakly supervised segmentation methods produce inferior results to fully supervised segmentation.

Pixel-level annotations allow fully supervised semantic segmentation to achieve reliability in learning the boundaries of objects and the relationship between their components. But, it is difficult to use image-level annotations to train segmentation networks because weakly labeled data only indicates the existence of objects of a certain class, and does not provide any information about their locations or boundaries.
Most weakly supervised methods using image-level annotations depend on localization maps obtained by a classification network~\cite{zhou2016learning} to bridge the gap between image-level  and pixel-level annotations. 
However, these localization maps focus only on the small discriminative parts of objects, without precise representations of their boundaries.
To bring the performance of these methods closer to that of fully supervised image segmentation means diverting the classifier from its primary task of discrimination between objects to discovering the relations between pixels.

We address this problem with FickleNet, which can generate a variety of localization maps from a single image using random combinations of hidden units in a convolutional neural network, as shown in Figure~\ref{fig_1}(a).
Starting with a feature map created by a generic classification network such as VGG-16~\cite{simonyan2014very}, FickleNet chooses hidden units at random for each sliding window position, which corresponds to each stride in the convolution operation, as shown in Figure~\ref{fig_1}(b). This process is simply realized by the dropout method~\cite{srivastava2014dropout}.
Selecting all the available hidden units in a sliding window position (the deterministic approach) tends to produce a smoothing effect that confuses foreground and background, which can result in both areas being activated or deactivated together. However, random selection of hidden units (the stochastic approach) produces regions of different shapes which can delineate objects more sharply.
Since the patterns of hidden units randomly selected by FickleNet include the shapes of the kernel of the dilated convolution with different dilation rates, FickleNet can be regarded as a generalization of dilated convolution, but FickleNet can potentially match objects of different scales and shapes using only a single network because it is not limited to a square array of hidden units, whereas dilated convolution requires networks with different dilation rates just to scale its kernel. 
%
%

The selection of random hidden units at each sliding window position is not an operation that is optimized at the CUDA level in common deep-learning frameworks such as PyTorch~\cite{paszke2017automatic}. 
Thus, a naive implementation of FickleNet, in which random hidden units are selected at each sliding window position and then convolved, would require a large number of iterative operations.
However, we can use the optimized convolution functions provided by deep-learning frameworks, if we expand the feature maps before making the random selection of hidden units.
The maps need to be expanded sufficiently to prevent successive sliding window positions from overlapping. We can then apply dropout in the spatial axis of the expanded feature maps, and perform a convolution operation with a stride equal to the kernel size.
This saves a significant amount of time without much increase in GPU memory usage, because the number of parameters to be back-propagated remains constant.

While many existing networks use stochastic regularization in their training process (e.g. Dropout ~\cite{srivastava2014dropout}), stochastic effects are usually excluded from the inference process.
However, our inference process contains random processes and thus produces a variety of localization maps.
The pixels that were allocated to a specific class with high scores in each localization map are discovered, and those pixels are aggregated into a single localization map.
The localization map obtained from FickleNet is utilized as pseudo-labels for the training of a segmentation network.


The main contributions of this paper can be summarized as follows:
\begin{itemize}
	\item[$\bullet$] We propose FickleNet, which is simply realized using the dropout method, that discovers the relationship between locations in an image and enlarges the regions activated by the classifier. 
	\item[$\bullet$] We introduce a method of expanding feature maps which makes our algorithm much faster, with only a small cost in GPU memory. 
	\item[$\bullet$] Our work achieves state-of-the-art performance on the Pascal VOC 2012 benchmark in both weakly supervised and semi-supervised settings.
\end{itemize}

\section{Related Work}
Weakly supervised semantic image segmentation methods substitute inexact annotations such scribbles, bounding boxes, or image-level annotations, for strong pixel-level annotations.
The methods of recent introduction have achieved successful results using annotations that provide location information such as scribbles or bounding boxes~\cite{tang2018regularized, dai2015boxsup}.
We now review some recently introduced weakly supervised approaches which use image-level annotations. 

A class activation map (CAM)~\cite{zhou2016learning} is a good starting-point for the classification of pixels from image-level annotations. A CAM discovers the contribution of each hidden unit in a neural net to the classification score, allowing the hidden units which make large contributions to be identified. 
However, a CAM tends to focus on the small discriminative region of a target object, which makes it unsuitable for training a semantic segmentation network.
Weakly supervised methods of recent introduction expand the regions activated by a CAM, operating on the image~(Section~\ref{image-level-processing}), on features~(Section~\ref{feature-level-preocessing}), or by growing the regions found by a CAM~(Section~\ref{region-growing}).

\subsection{Image-level Processing}\label{image-level-processing}
Image-level hiding and erasure have been proposed~\cite{singh2017hide,wei2017object,li2018tell} as ways of preventing a classifier from focusing exclusively on the discriminative parts of objects. Hide-and-Seek~\cite{singh2017hide} hides random regions of a training image, forcing the classification network to seek other parts of the object. However, the process of hiding random regions does not consider the semantics and sizes of objects.
Adversarial Erasing~\cite{wei2017object} starts with
a single small region in the object, and then drives the classification network to discover a sequence of new and complement any object regions by erasing the regions that have already been found. Although it can progressively expand regions belonging to an object, it requires multiple classification networks to perform the repetitive classification and erasure steps. 
The Guided Attention Inference Network (GAIN)~\cite{li2018tell} has a CAM which is trained to erase regions in a way that deliberately confuses the classifier.
This CAM has to be large enough to cover an entire object. 
However, the classifier mainly reacts to high activation, and so it can become confused if an object's only discriminative parts are erased.
\subsection{Feature-level Processing}\label{feature-level-preocessing}
Feature-level processing can be used to expand the regions activated by a CAM. Adversarial complementary learning~\cite{zhang2018adversarial} and two-phase learning~\cite{kim2017two} use a classifier to identify the discriminative parts of an object and erase them based on features. A second classifier then is trained to find the complementary parts of the object from those erased features.
This is an efficient technique which operates at a relatively high level. 
However, it has a similar drawback to image-level erasure, in that a second classifier and training step are essential for those methods, which may cause a suboptimal performance.
In addition, features whose discriminative parts are erased can confuse the second classifier, which may not be correctly trained.
Pyramid Grad-CAM~\cite{lee2018robust} considers multi-layer features for multi-scale context.

Multi-dilated convolution (MDC)~\cite{wei2018revisiting} uses several convolutional blocks, dilated at different rates, within a generic classification network, and aggregates CAMs obtained from each block in a process that resembles ensemble learning. The different-sized receptive fields produced by different dilation rates can be shown to capture different patterns, but MDC requires a separate training procedure for each dilation rate, and its limitation to integer dilation rates (e.g. 1, 3, 6, 9) means that only a limited number of ensembles is possible.
In addition, the receptive field produced by a standard dilated convolution is square with a fixed size, so that MDC tends to identify false positive regions.

\subsection{Region Growing}\label{region-growing}
Region growing can be used to expand the localization map produced by a CAM, which initially identifies just the small discriminative part of an object.
AffinityNet~\cite{ahn2018learning} learns pixel-level semantic affinities, which identify pixels belonging to the same object, under the supervision of an initial CAM, and then expands the initial CAM by a random walk with the transition matrix computed from semantic affinities. However, the learning of semantic affinities requires an additional network, and the outcome depends heavily on the quality of the CAM.
Seed, Expand, and Constrain (SEC)~\cite{kolesnikov2016seed} uses a new type of loss function to expand the localization map and constrain it to object boundaries using a conditional random field (CRF)~\cite{krahenbuhl2011efficient}. Deep seeded region growing (DSRG)~\cite{huang2018weakly} refines initial localization maps during the training of its segmentation network, so that DSRG does not require additional networks to grow regions.
The seeds for region growing are obtained from a CAM, and if these seeds only come from the discriminative parts of objects, it is difficult to grow regions into non-discriminative parts. We therefore utilize as a segmentation network with the localization maps produced by FickleNet.
\begin{figure}[t]
  \centering
  \includegraphics[width=1.0\linewidth]{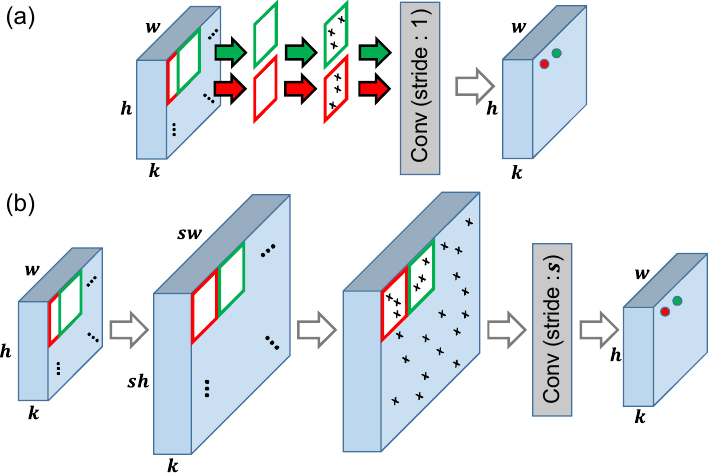}
  \caption{\label{fig_expand} 
  (a) Naive implementation of FickleNet, which requires a dropout and convolution function call at each sliding window position (the red and green boxes). (b) Implementation using map expansion: convolution is now performed once with a stride of $s$. The input feature map is expanded so that successive sliding kernels (the red and green boxes) do not overlap.}
  \vspace*{-10pt}
\end{figure}

\section{Proposed Method}
Our procedure has the following steps: FickleNet, which uses stochastic selection of hidden units, is trained for multi-class classification. It then generates localization maps of training images. Finally, the localization maps are used as pseudo-labels to train a segmentation network. 
We denote the sort of feature map typically obtained from a standard deep neural network as $x \in \mathbb{R}^{k \times h \times w}$, where $w$ and $h$ are the width and the height of each of $k$ channels, respectively. The procedures for training FickleNet and generating localization maps are shown as Algorithm~\ref{Procedure}.
\begin{algorithm}
\footnotesize
\KwIn{Image $I$, ground-truth label $c$, dropout rate $p$}
\KwOut{Classification score $S$ and localization maps $M$}
$x$ = Forward($I$) until conv5 layer; \\
\textbf{Stochastic hidden unit selection:}\hspace*{\fill} Sec.~\ref{SDC}\\
\qquad $x^{\text{expand}}$ = Expand(x);\hspace*{\fill} Sec.~\ref{Expand_Trick}\\
\qquad $x^{\text{expand}}_p$ = Center-fixed spatial dropout($x^{\text{expand}}$, $p$);\hspace*{\fill} Sec.~\ref{dropout}\\
\qquad $S$ = Classifier($x^{\text{expand}}_p$);\hspace*{\fill} Sec.~\ref{training}\\
\textbf{Training Classifier:}\\
\qquad Update network by $L$=SigmoidCrossEntropy($S$, $c$)\\
\textbf{Inference CAMs:}\hspace*{\fill} Sec.~\ref{CAMextract}\\  
\qquad For different random selections $i~(1 \leq i \leq N)$:\\
\qquad\qquad $M^c[i]$ = Grad-CAM$(x, S^c)$; \hspace*{\fill} Sec.~\ref{gradcam}\\
\qquad $M^c$ = Aggregate($M^c[i]$);\hspace*{\fill} Sec.~\ref{aggregate}\\
\caption{Training and Inference Procedure}\label{Procedure}
\end{algorithm}

\subsection{Stochastic Hidden Unit Selection}\label{SDC}
Stochastic hidden unit selection is used in FickleNet to discover relations between parts of objects by exploring the classification score computed from the randomly selected pairs of hidden units, with the aim of associating a non-discriminative part of an object with a discriminative part of the same object. This process is realized by applying spatial dropout~\cite{srivastava2014dropout} to the feature $x$ at each sliding window position, as shown in Figure~\ref{fig_expand}(a).
This differs from the standard dropout technique, which only samples hidden units in the feature maps once in each forward pass, and thus hidden units which are not sampled cannot contribute to the class scores.
Our method samples hidden units at each sliding window position, which means that a hidden unit may be activated at some window positions and dropped at others.

\begin{figure}[t]
  \centering
  \includegraphics[width=1.0\linewidth]{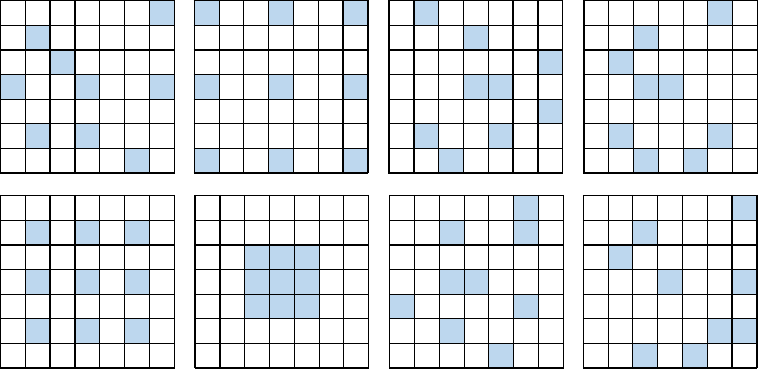}
  \caption{\label{fig_dilation} Examples of the selection of 9 hidden units (marked as blue) from a 7 $\times$ 7 kernel. Channels are not shown for simplicity. The shapes of those selected hidden units sometimes contain the shape of kernel of convolution with different dilation rates.}
  
\end{figure}

This method of selecting hidden units can generate receptive fields of many different shapes and sizes, as shown in Figure~\ref{fig_dilation}.
Some of these fields are likely to be similar to those produced by a standard dilated convolution; thus the results produced by this technique can be expected to contain those produced by standard dilated convolution at various rates.
This selection process can be simply and efficiently realized by the expansion technique described in Section~\ref{Expand_Trick} with a method which we call center-preserving dropout, which is described in Section~\ref{dropout}.
\subsubsection{Feature Map Expansion} \label{Expand_Trick}
As our method needs to sample new combinations in each sliding window position, we cannot directly utilize the CUDA-level optimized convolution functions provided by popular deep learning frameworks such as PyTorch~\cite{paszke2017automatic}. If we were to implement our method naively, as shown in Figure~\ref{fig_expand}(a), we would have to call the convolution function and the dropout function in $w \times h$ times in each forward pass. By expanding the feature map, we reduce this to a single call to each function during each forward pass.

Figure~\ref{fig_expand}(b) shows how we expand the input feature maps so that no sliding window positions overlap.
Before expanding the feature map, we apply zero padding on $x$ so that the size of the final output is equal to that of the input.
The size of the feature map after zero padding becomes ${k \times (h + s - 1) \times (w + s - 1)}$, where $s$ is the size of the convolution kernel.
We expand the zero-padded feature map so that successive sliding window positions do not overlap, and the size of the expanded feature map $x^{\text{expand}}$ is $k \times (sh) \times (sw)$. 
We then select hidden units on $x^{\text{expand}}$ using the center-preserving dropout technique explained in Section~\ref{dropout}.
Although the expanded feature map requires more GPU memory, the number of parameters to be trained remains constant, and so the load on the GPU does not increase significantly.

\subsubsection{Center-preserving Spatial Dropout} \label{dropout}
We realize stochastic hidden unit selection by applying the dropout method~\cite{srivastava2014dropout} to spatial locations. We can achieve the same results as the naive implementation by applying dropout only once to the expanded feature map $x^{\text{expand}}$.
Note that dropout is applied uniformly across all channels.
We do not drop the center of the kernel of each sliding window position, so that relationships between kernel center and other locations in each stride can be found. After spatial dropout with a rate of $p$, we denote the modified feature map as $x^{\text{expand}}_p$. 
While dropout is usually only employed during training, we apply it to both training and inference.
\\[-1.95em]
\subsubsection{Classification} \label{training}
In order to obtain classification scores, convolution with kernel of size $s$ and stride $s$ are applied to the dropped feature map $x^{\text{expand}}_p$. We then obtain an output feature map of size $c \times w \times h$, where $c$ is the number of object classes. 
By applying global average pooling and a sigmoid function to this map, we obtain a classification score $S$. We then update FickleNet using the sigmoid cross-entropy loss function, which is widely used for multi-label classification. 

\subsection{Inference Localization Map} \label{CAMextract}
We can now obtain various classification scores from a single image, which correspond to randomly selected combinations of hidden units, and each random selection generates a various localization map. Section~\ref{gradcam} describes how to obtain a localization map from each random selection, and Section~\ref{aggregate} describes how the maps from the random selections are aggregated into a single localization map.
\\[-1.95em]\subsubsection{Grad-CAM} \label{gradcam}
We use gradient based CAM (Grad-CAM)~\cite{selvaraju2017grad}, which is a generalization of class activation map (CAM)~\cite{zhou2016learning}, to obtain localization maps. Grad-CAM discovers the class specific contribution of each hidden unit to the classification score from gradient flows. We compute the gradients of the target class score with respect to $x$, which is the feature map before expansion, and then sum the feature maps along the channel axis, weighted by these gradients.
We can express Grad-CAM for each target class $c$ as follows:
\begin{equation}\label{gradcam1}
    \text{Grad-CAM}^{\text{c}} = \text{ReLU}(\sum_{k} x_k \times ~\frac{\partial S^{c}}{\partial x_k}),
\end{equation}
where $x_k \in \mathbb{R}^{w \times h}$ is the $k^{th}$ channel of the feature map $x$, and $S^{c}$ is the classification score of class $c$.

\subsubsection{Aggregate Localization Map} \label{aggregate}
FickleNet allows many localization maps to be constructed from a single image, because different combinations of hidden units are used to compute classification scores at each random selection. We construct $N$ different localization maps from a single image and aggregate them into a single localization map. Let $M[i]~(1\leq i \leq N)$ denote the localization map constructed from the $i^{th}$ random selection. 
We aggregate the $N$ localization maps so that a pixel located at $u$ in the aggregated map is allocated to class $c$ if the activation score for class $c$ in any $M[i]$ at $u$ is higher than a threshold $\theta$. Pixels which are not allocated to any class are ignored during training.
If there is a pixel assigned to multiple classes, we examine its class score in a map averaged over the $N$ maps and assign the pixel to the class with the highest score in the average map.


\subsection{Training the Segmentation Network} \label{segtraining}
The localization map, whose construction was described in Sections~\ref{SDC} and~\ref{CAMextract}, provides pseudo-labels to train a semantic image segmentation network.
We use the same background cues as DSRG~\cite{huang2018weakly}. We feed the generated localization maps from FickleNet to DSRG as the seed cues for weakly supervised segmentation.

For semi-supervised learning we introduce an additional loss derived from data fully annotated by a person. Let $\mathcal{C}$ be the set of classes that are present in the image. We train a segmentation network with the following loss function:
\begin{align}\label{semi_loss_total}
    L = L_{\text{seed}} + L_{\text{boundary}} + \alpha L_{\text{full}},
\end{align}
where $L_{\text{seed}}$ and $L_{\text{boundary}}$ respectively are the balanced seeding loss and boundary loss used in DSRG~\cite{huang2018weakly}, and 
\begin{align}\label{semi_loss_full}
    L_{\text{full}} = - \frac{1}{\sum\limits_{c \in \mathcal{C}} |F_c|} \sum_{c \in \mathcal{C}}\sum_{u \in F_c} \log H_{u, c},
\end{align}
where $H_{u, c}$ is the probability of an entry of class $c$ at location $u$ in the segmentation map $H$, and $F_c$ is the ground-truth mask.

\section{Experiments}
\subsection{Experimental Setup}\label{impledetail}
\noindent\textbf{Dataset:} We conducted experiments on the PASCAL VOC 2012 image segmentation benchmark~\cite{everingham2010pascal}, which contains 21 object classes, including one background class. Using the same protocol as other work on weakly supervised semantic segmentation, we trained our network using augmented 10,582 training images with image-level annotations. We report mean intersection-over-union (mIoU) for 1,449 validation images and 1,456 test images.
The results for the test images were obtained on the official PASCAL VOC evaluation server. 

\noindent\textbf{Network details:} 
FickleNet is based on the VGG-16 network~\cite{simonyan2014very}, pre-trained using the Imagenet~\cite{deng2009imagenet} dataset. The VGG-16 network was modified by removing all fully-connected layers and the last pooling layer, and we replaced the convolution layers of the last block with dilated convolutions with a rate of $2$.
We set the kernel size $s$ and the dropout rate $p$ to 9 and 0.9 respectively.
Segmentation is performed by DSRG~\cite{huang2018weakly} based on Deeplab-CRF-LargeFOV~\cite{chen2014semantic}.

\noindent\textbf{Experimental details:}
We trained FickleNet using a mini-batch size to 10. We cropped the training images to $321 \times 321$ pixels at random locations, so that the size of feature map $x$ becomes $512 \times 41 \times 41$. The initial learning rate was set to 0.001 and halved every 10 epochs. We used the Adam optimizer~\cite{kingma2014adam} with its default settings. During segmentation training, we use the same settings as DSRG~\cite{huang2018weakly}. We set the number of different localization maps $N$ for each image to $200$ and the threshold $\theta$ to 0.35. We set $\alpha$ to 2 for semi-supervised learning.

\noindent\textbf{Reproducibility:} PyTorch~\cite{paszke2017automatic} was used for training FickleNet and conducting localization maps, and we used the Caffe deep learning framework~\cite{jia2014caffe} in the segmentation step. 
All the experiments were performed on an NVIDIA TITAN Xp GPU. We will soon make both our code and the trained models publicly available.

\begin{table}[htbp]\setlength\heavyrulewidth{0.25ex}
  \centering
  \caption{Comparison of weakly supervised semantic segmentation
methods on VOC 2012 validation and test image sets. The methods listed here use DeepLab-VGG16 for segmentation.}
    \begin{tabular}{lccc}
    \Xhline{1pt}\\[-0.95em]
    Methods & Training & \textit{val} & \textit{test} \\
    \hline\hline 
    \\[-0.9em]
    \multicolumn{4}{l}{Supervision: Image-level and additional annotations} \\
    $\text{MIL-seg}_{\text{~~CVPR '15}}$~\cite{pinheiro2015image} & 700K  & 42.0    & 40.6 \\
    $\text{STC}_{\text{~~TPAMI '17}}$~\cite{wei2017stc}   & 50K   & 49.8  & 51.2 \\
    $\text{TransferNet}_{\text{~~CVPR '16}}$~\cite{hong2016learning} & 70K   & 52.1  & 51.2 \\
    $\text{CrawlSeg}_{\text{~~CVPR '17}}$~\cite{hong2017weakly}& 970K  & 58.1  & 58.7 
      \\
      $\text{AISI}_{\text{~~ECCV '18}}$~\cite{hu2018associating}& 11K  & 61.3  & 62.1 
      \\\\[-0.9em]
    
    \hline
    \\[-0.9em]
    \multicolumn{4}{l}{Supervision: Image-level annotations only}\\
    $\text{SEC}_{\text{~~ECCV '16}}$~\cite{kolesnikov2016seed} & 10K   & 50.7  & 51.1 \\
    $\text{CBTS-cues}_{\text{~~CVPR '17}}$~\cite{roy2017combining} & 10K   & 52.8  & 53.7 \\
    $\text{TPL}_{\text{~~ICCV '17}}$~\cite{kim2017two} & 10K   & 53.1  & 53.8 \\
    $\text{AE\_PSL}_{\text{~~CVPR '17}}$~\cite{wei2017object} & 10K   & 55.0    & 55.7 \\
    $\text{DCSP}_{\text{~~BMVC '17}}$~\cite{chaudhry2017discovering} & 10K & 58.6 & 59.2\\
    $\text{MEFF}_{\text{~~CVPR '18}}$~\cite{ge2018multi} & 10K & - & 55.6\\
    $\text{GAIN}_{\text{~~CVPR '18}}$~\cite{li2018tell} & 10K   & 55.3  & 56.8 \\
    $\text{MCOF}_{\text{~~CVPR '18}}$~\cite{wang2018weakly} & 10K   & 56.2  & 57.6 \\
    $\text{AffinityNet}_{\text{~~CVPR '18}}$~\cite{ahn2018learning} & 10K   & 58.4  & 60.5 \\
    $\text{DSRG}_{\text{~~CVPR '18}}$~\cite{huang2018weakly} & 10K   & 59.0    & 60.4 \\
    $\text{MDC}_{\text{~~CVPR '18}}$~\cite{wei2018revisiting}& 10K   & 60.4  & 60.8 \\
    FickleNet (Ours) & 10K   &  \textbf{61.2}  &  \textbf{61.9}\\
    \Xhline{1pt}
    \end{tabular}%
  \label{sotacomparison}%
  \vspace*{-6pt}
\end{table}%
\begin{table}[htbp]
  \centering
  \caption{Comparison of weakly supervised semantic segmentation
methods on VOC 2012 validation and test image sets. The methods listed here use ResNet-based DeepLab for segmentation.}
    \begin{tabular}{lccc}
    \Xhline{1pt}
        \\[-0.9em]
    Methods & Backbone & ~\textit{val} ~  & ~\textit{test} ~\\
    \hline\hline    \\[-0.9em]
    MCOF~\cite{wang2018weakly}& ~~ResNet 101 ~ & ~60.3~ & 61.2~\\
    DCSP~\cite{chaudhry2017discovering}& ResNet 101~&  ~ 60.8  ~  &  ~61.9 ~\\
    DSRG~\cite{huang2018weakly}& ResNet 101~&  ~ 61.4  ~  &  ~63.2 ~\\
    AffinityNet~\cite{ahn2018learning}& ResNet 38~ &  ~ 61.7  ~  & ~63.7 ~\\
    FickleNet (ours) &~ResNet 101~ & ~  \textbf{64.9}  ~  &~\textbf{65.3}  ~\\
    \Xhline{1pt}
    \end{tabular}%
  \label{resnets}%
  \vspace*{-14pt}
\end{table}%
\subsection{Comparison to the State of the Art}
\noindent\textbf{Weakly supervised segmentation:} We compared our method with other recently introduced weakly supervised semantic segmentation methods with various levels of supervision. Table~\ref{sotacomparison} shows results on PASCAL VOC 2012 images.
Our method outperformed others which provide the same level of supervision through image-level annotations, achieving mIoU values of $61.2$ and $61.9$ for validation and test images respectively. This represents a 2.2\% and 1.5\% improvement respectively on validation and test images, when compared to DSRG, which is our backbone network. The performance of FickleNet was 91\% of that of DeepLab~\cite{chen2014semantic} trained with fully annotated data, which achieved an mIoU of 67.6 on validation images. 
Note that we do not need additional training steps or additional networks, in contrast to many other recent techniques, such as AffinityNet~\cite{ahn2018learning}, which requires an additional network for learning semantic affinities, or AE-PSL~\cite{wei2017object} and MDC~\cite{wei2018revisiting}, which require several training steps. Table~\ref{resnets} shows result on PASCAL VOC 2012 images with a ResNet-based segmentation network. We achieved mIoU values of $64.9$ and $65.3$ for validation and test images respectively using DeepLab-v2-ResNet101. This represents a 3.5\% and 2.1\% improvement, respectively, on validation and test images, when compared to DSRG. AffinityNet~\cite{ahn2018learning} uses ResNet-38 based network~\cite{wu2016wider}, which is more powerful than ResNet-101. Note that FickleNet, which is used for conducting localization maps, is still based on VGG-16. 
\begin{table}[htbp]
  \centering  \caption{Comparison of semi-supervised semantic segmentation methods on VOC 2012 validation sets. We also give the performances of DeepLab using 1.4K and 10.6K strongly annotated data.}
  \resizebox{0.48\textwidth}{!}{
    \begin{tabular}{lcc}
    \Xhline{1pt}\\[-0.95em]
    Methods & Training Set &mIoU    \\
    \hline\hline\\[-0.95em]
    DeepLab~\cite{chen2014semantic} & 1.4K strong & 62.5\\
    \hline\\[-0.95em]
    WSSL~\cite{papandreou2015weakly}  & 1.4K strong + 9K weak & 64.6  \\
    GAIN~\cite{li2018tell}  & 1.4K strong + 9K weak & 60.5  \\
    MDC~\cite{wei2018revisiting}  & 1.4K strong + 9K weak & 65.7 \\
    DSRG ~\cite{huang2018weakly} (baseline) & 1.4K strong + 9K weak& 64.3\\
    FickleNet (ours)& 1.4K strong + 9K weak &   \textbf{65.8}   \\
    \hline \\[-0.95em]
    DeepLab~\cite{chen2014semantic} & 10.6K strong & 67.6  \\
    \Xhline{1pt}
    \end{tabular}%
  \label{tabsemi}}%
    \vspace*{-4pt}

\end{table}%
\begin{table}[htbp]
  \centering
  \caption{Run time and GPU memory usage for training and CAM extraction without and with map expansion.}
    \begin{tabular}{lccc}
    
    \Xhline{1pt}
    \\[-0.9em]
      Methods   & Training & CAM Extract & GPU Usage\\
      \hline
    \hline
    \\[-0.9em]
    Naive &   20 sec/iter    & 2.98 sec/img &  8.4 GB\\
    Expansion&  1.3 sec/iter & 0.21 sec/img & 10.1 GB\\
    \Xhline{1pt}
    \end{tabular}%
  \label{runtime_table}%
    \vspace*{-4pt}
\end{table}%
\begin{table}[htbp]
\renewcommand{\arraystretch}{0.9}
  \centering
  \caption{Comparison of mIoU scores using different dropout rates ($p$) on PASCAL VOC 2012 validation images.}
    \begin{tabular}{lcc}
    \Xhline{1pt}\\[-0.85em]
    Methods & Dropout Rate ($p$) & mIoU \\
    \hline\hline\\[-0.88em]
    Deterministic ~~~& 0.0 & 56.3 \\
    \hline\\[-0.89em]
    \multicolumn{1}{l}{\multirow{2}[0]{*}{General Dropout}} &  0.5 & 45.6\\
    \multicolumn{1}{l}{} & 0.9 &  49.1 \\
    \hline\\[-0.97em]
    \multicolumn{1}{l}{\multirow{4}[0]{*}{FickleNet}} &0.3 & 58.8 \\
    \multicolumn{1}{l}{} &  0.5 &  59.4\\
    \multicolumn{1}{l}{} &  0.7 &  60.0 \\
    \multicolumn{1}{l}{} &  0.9 & \textbf{61.2} \\
    \Xhline{1pt}
    \end{tabular}%
  \label{full_result_rate}%
\vspace*{-8pt}
\end{table}%

Our method also significantly outperforms methods based on additional supervision except AISI~\cite{hu2018associating}. These methods include TransferNet~\cite{hong2016learning}, which was trained on pixel-level annotations of 60 classes (not Pascal VOC classes) of COCO~\cite{lin2014microsoft} images, and CrawlSeg~\cite{hong2017weakly}, which was provided with a very large number of unlabeled YouTube videos. AISI~\cite{hu2018associating} utilized salient instance detector~\cite{fan2017s} which is trained using well-annotated instance-level annotations. Note that instance-level annotation is one of the most difficult annotations to obtain.

\begin{figure*}[t]
  \centering
  \includegraphics[width=0.93\linewidth]{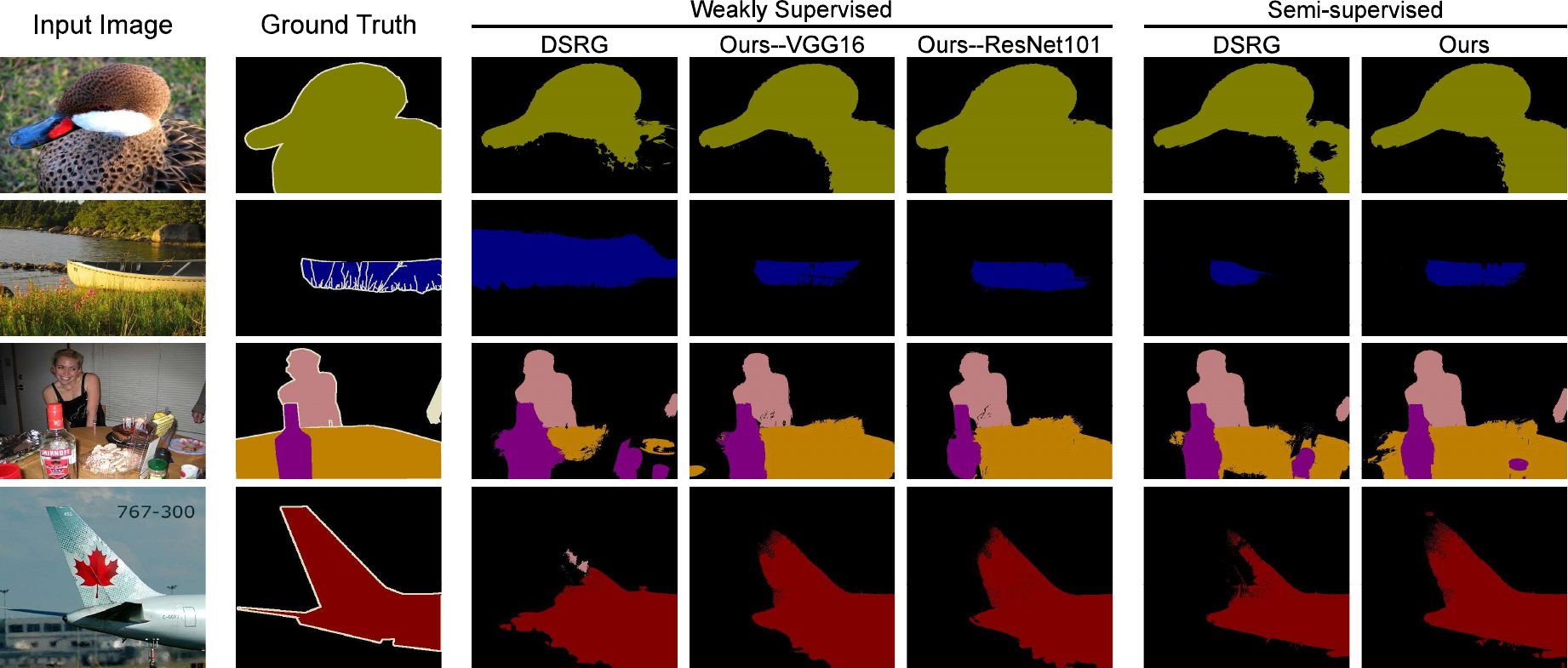}
  \caption{\label{seg_ex} Examples of predicted segmentation masks for Pascal VOC 2012 validation images in weakly and semi-supervised manner.}
\end{figure*}

Figure~\ref{seg_ex} shows qualitative results of predicted segmentation masks, in FickleNet and DSRG, which is our backbone network. The supervision provided by FickleNet produces larger and more accurate regions of a target object than that used in DSRG, allowing the segmentation network to consider a wider range of target objects.
 Thus, the segmentation network trained with localization maps generated by FickleNet produces more accurate results than DSRG in that FickleNet can make fewer false positives and cover larger regions of a target object.

\begin{figure*}[t]
  \centering
  \includegraphics[width=1.0\linewidth]{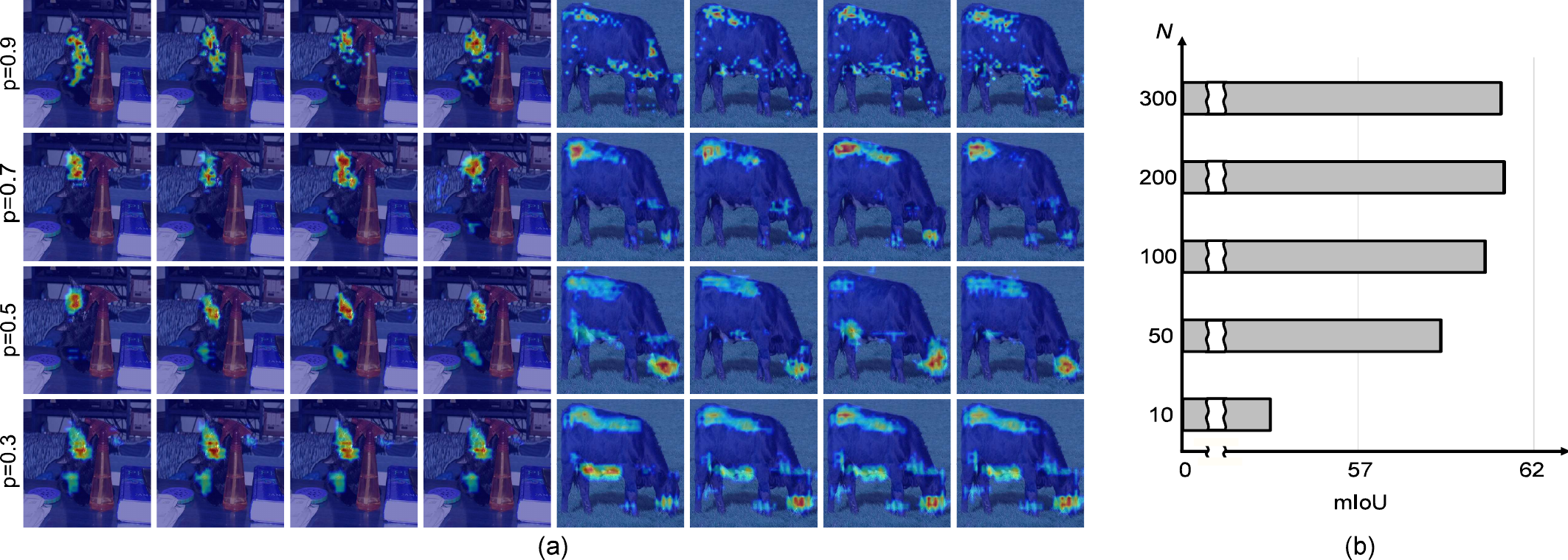}
  \caption{\label{cam_instances} (a) Localization maps from each random selection of hidden unit with different dropout rates $p$. (b) Performance on Pascal VOC 2012 validation images for different $N$.}

\end{figure*}

\noindent\textbf{Semi-supervised segmentation:} Table~\ref{tabsemi} shows that the mIoU of 65.8 produced by our method, trained on only 13.8\% of images with pixel-level annotations in the PASCAL dataset, was 97.3\% of that of Deeplab, which is trained with fully annotated data.
The performance of FickleNet on validation images was 1.5\% better than that of DSRG which is our baseline network. 
Note that GAIN shows lower performance than Deeplab, trained on only 1.4K fully annotated data. GAIN uses pixel-level annotations for the training of a classifier, rather than a segmentation network so that pixel-level ground-truth indirectly affects the training of the segmentation network.
Figure~\ref{seg_ex} shows examples of segmentation maps from DSRG and FickleNet, which demonstrate that our system is able to operate satisfactorily in a semi-supervised manner.

\begin{figure*}[t]
  \centering
  \includegraphics[width=1.0\linewidth]{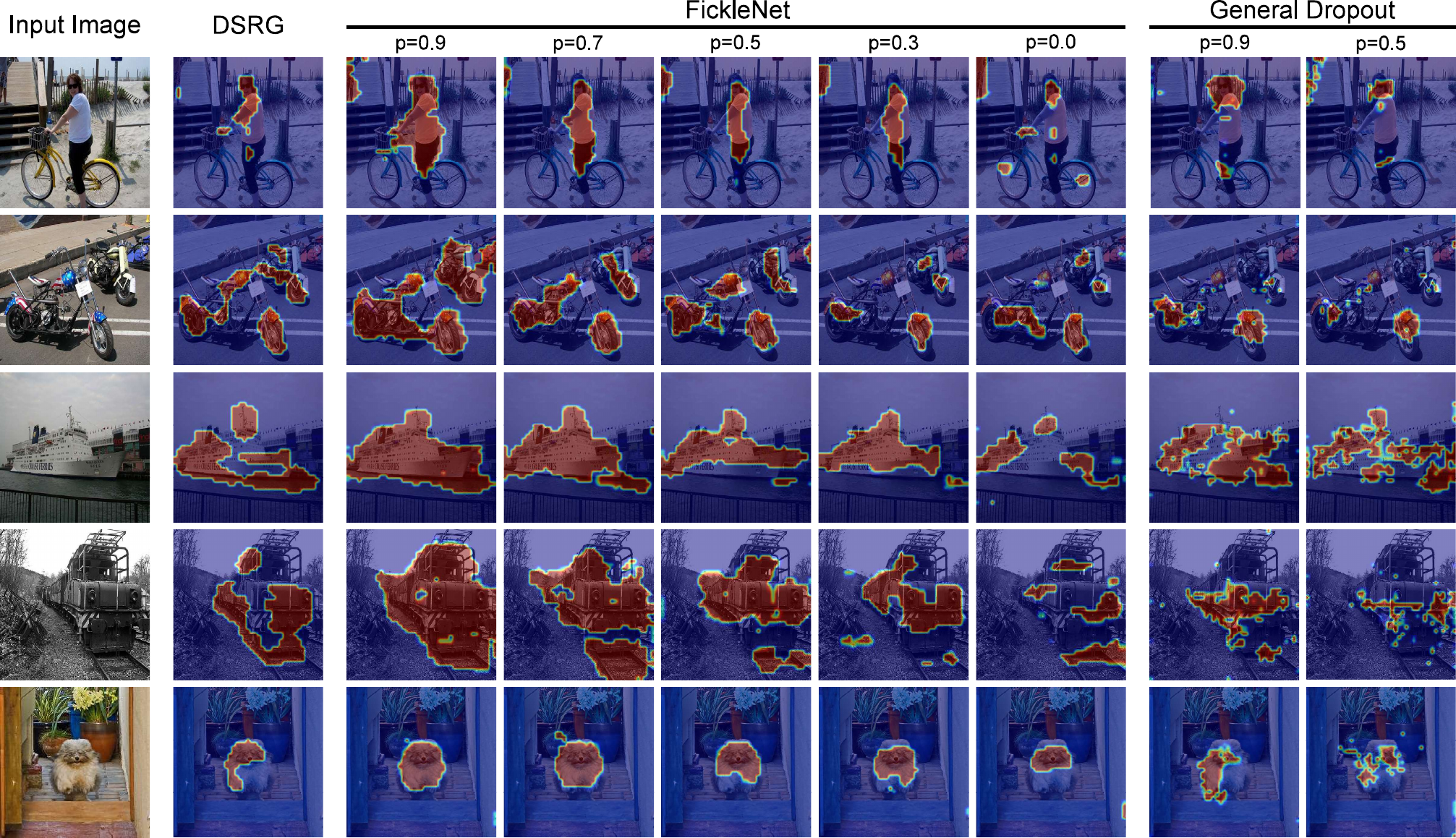}
  \caption{\label{cam_samples} Localization maps from DSRG and FickleNet, with various dropout rates (p = 0 denotes a deterministic network), and from the general dropout method. Localization maps of DSRG (the $2^{\text{nd}}$ column) were visualized using the publicly available DSRG localization cue.}
   \vspace*{-10pt}
\end{figure*}
\subsection{Ablation studies}

%
%
\subsubsection{Effects of the Map Expansion Technique}
In order to show the effect of the map expansion technique presented in Section~\ref{Expand_Trick}, we compare runtime and GPU usage of a naive implementation of FickleNet (Fig.~\ref{fig_expand}(a)) with that of an implementation of FickleNet with map expansion (Fig.~\ref{fig_expand}(b)). Table~\ref{runtime_table} shows that training and CAM extraction times are reduced factors of 15.4 and 14.2 respectively, at a cost of 12\% in GPU memory usage.\\[-1.95em]
\subsubsection{Analysis of Iterative Inference}
We compare mIoU scores on PASCAL VOC 2012 validation images with different numbers of localization maps $N$ from a single image. Figure~\ref{cam_instances}(b) shows that the mIoU increases with the number of maps $N$.
We interpret this as meaning that additional random selection identifies more regions of a target object, so that larger regions of that object are represented by the aggregated localization map. If $N$ is greater than 200, the mIoU converges to 61.2. Examples of different CAMs obtained from a single image are shown in Figure~\ref{cam_instances}(a). 

\begin{table}[thbp]
  \centering
  \caption{Standard deviation of mIoU, recall, precision (direct measures).}
  \resizebox{0.42\textwidth}{!}{
    \begin{tabular}{ccccc}
    \Xhline{1pt}
    N     & 10    & 100   & 200   & 300 \\
    \hline\hline
    std (mIoU, $10^{-3}$) & 21 & 14 & 6.8 & 4.8 \\
    std (recall, $10^{-5}$) & 22.4 & 14.8 & 6.72 & 3.41 \\
    std (prec, $10^{-5}$) & 27.7 & 12.3 & 8.77 & 9.99 \\
    \Xhline{1pt}
    \end{tabular}%
    }
  \label{std}%
\end{table}%

\begin{table}[thbp]
  \centering
  \caption{Effectiveness of each step. $\mathcal{G}-$ general dropout, $\mathcal{S}-$ stochastic selection, $\mathcal{D}-$ deterministic approach.}
    \resizebox{0.43\textwidth}{!}{

    \begin{tabular}{c|cccccc}
    \Xhline{1pt}\\[-0.97em]
    Training & $\mathcal{G}$  & $\mathcal{G}$ & $\mathcal{G}$ & $\mathcal{S}$ & $\mathcal{S}$& $\mathcal{D}$ \\
    Inference & $\mathcal{G}$     & $\mathcal{S}$ & $\mathcal{D}$ & $\mathcal{S}$ & $\mathcal{D}$& $\mathcal{D}$ \\
    \hline\hline
    mIoU  & 49.1  &    55.5   & 57.1  & \textbf{61.2}  & 59.6  & 59.0 \\
    \Xhline{1pt}
    \end{tabular}%
    }
    
    \vspace*{-10pt}
  \label{stage_eval}%
\end{table}%
We make two observations to support the stability of iterative stochastic selection: 1) The segmentation performances converge as $N$ increases (Figure~\ref{cam_instances}(b)). 2) When we perform feature selection 5 times with various values of $N$, the standard deviation of mIoU, recall, and precision, which is the direct measures from the seeds, are very small and drop further as $N$ increases (Table~\ref{std}).

\subsubsection{Analysis of Dropout}
\textbf{Effects of dropout rate:} We analyzed the effects of the dropout rate used by FickleNet.
Figure~\ref{cam_samples} shows that a dropout rate $p$ of 0.9 allows FickleNet to cover larger regions of the target object than DSRG, which uses the localization maps from deterministic classifiers. Higher dropout rates also lead to more widely activated localization maps, because it becomes more likely that the discriminative part of an object will be dropped, leaving the non-discriminative parts of the object to be considered for classification.
Conversely, if the dropout rate is low, the discriminative parts of objects are unlikely to be dropped, and they usually suffice for classification; so the classifier is unlikely to activate non-discriminative parts. As shown in Figure~\ref{cam_instances}(a), FickleNet with a low dropout rate tends to activate only the discriminative part of objects, even though random sampling produces many patterns of hidden units. Higher dropout rates result in more randomness in the activated patterns so that different non-discriminative parts of an object are more likely to be considered for each random selection. This effect is also reflected in the quantitative results shown in Table~\ref{full_result_rate}.


\noindent\textbf{Comparison to general dropout:} We compared FickleNet with a network created using a general dropout method rather than the hidden unit selection. Figure~\ref{cam_samples} shows that localization maps from the network created with general dropout tend to show noisy activation: hidden units which are not sampled cannot contribute to the class score during a forward pass, which means these dropped units do not contribute to the localization map. Note that a hidden unit in FickleNet may be activated at some window positions and dropped at others so that every hidden unit is able to affect the classification score. 
In Table~\ref{full_result_rate}, a segmentation network trained with localization maps from the network with general dropout shows inferior results to FickleNet. 

\noindent\textbf{Effectiveness of each step:} Table~\ref{stage_eval} shows results obtained using several combinations of general dropout ($\mathcal{G}$), stochastic selection ($\mathcal{S}$), and the deterministic approach ($\mathcal{D}$) for training and inference. As expected, ``train $\mathcal{S}$ + infer $\mathcal{D}$" is better than ``train $\mathcal{D}$ + infer $\mathcal{D}$", because stochastic selection lets the network consider the non-discriminative part, but the best mIoU is obtained by ``train $\mathcal{S}$ + infer $\mathcal{S}$".


%
%


\section{Conclusions}
We have addressed the problem of semantic image segmentation using only image-level annotations. By choosing features at random during both training and inference, we obtain many different localization maps from a single image, and then aggregate those maps into a single localization map. This map contains regions corresponding to parts of objects which are both larger and more consistent than those on a map produced by an equivalent deterministic technique. Our method can be implemented efficiently using operations readily available on a GPU by expanding the feature maps to avoid overlaps between the sliding kernels used during convolution. We show that the results produced by FickleNet on both weakly supervised and semi-supervised segmentation are better than those produced by other state-of-the-art approaches.

\setcounter{section}{0}
\renewcommand\thesection{\Alph{section}}
\renewcommand\thesubsection{\thesection.\arabic{subsection}}

{\small
\bibliographystyle{ieee}
\bibliography{egbib}
}

\newpage
\section{Appendix}




\subsection{Additional Results}

\noindent\textbf{Effects of conditional random field:}
Table~\ref{without_crf} shows results on PASCAL VOC 2012 validation images with and without the conditional random field (CRF), which is popularly used as a post-processing method.
The authors of DSRG~\cite{huang2018weakly} have not made their trained model publicly available; thus, we retrained DSRG to obtain the result without CRF. 
Note that the reported mIoU of DSRG with CRF is 59.0, which shows little difference with the result that we reproduced (58.9). 
FickleNet outperforms recent methods both with and without CRF and shows the least difference between with and without CRF.


\noindent\textbf{Per-class Results:} Table~\ref{class-specific-results} shows the per-class mIoU of DSRG, FickleNet, and some other settings of FickleNet. It outperforms DSRG in 17 classes among 21 classes.

\begin{figure*}[t]
  \centering
  \includegraphics[width=0.86\linewidth]{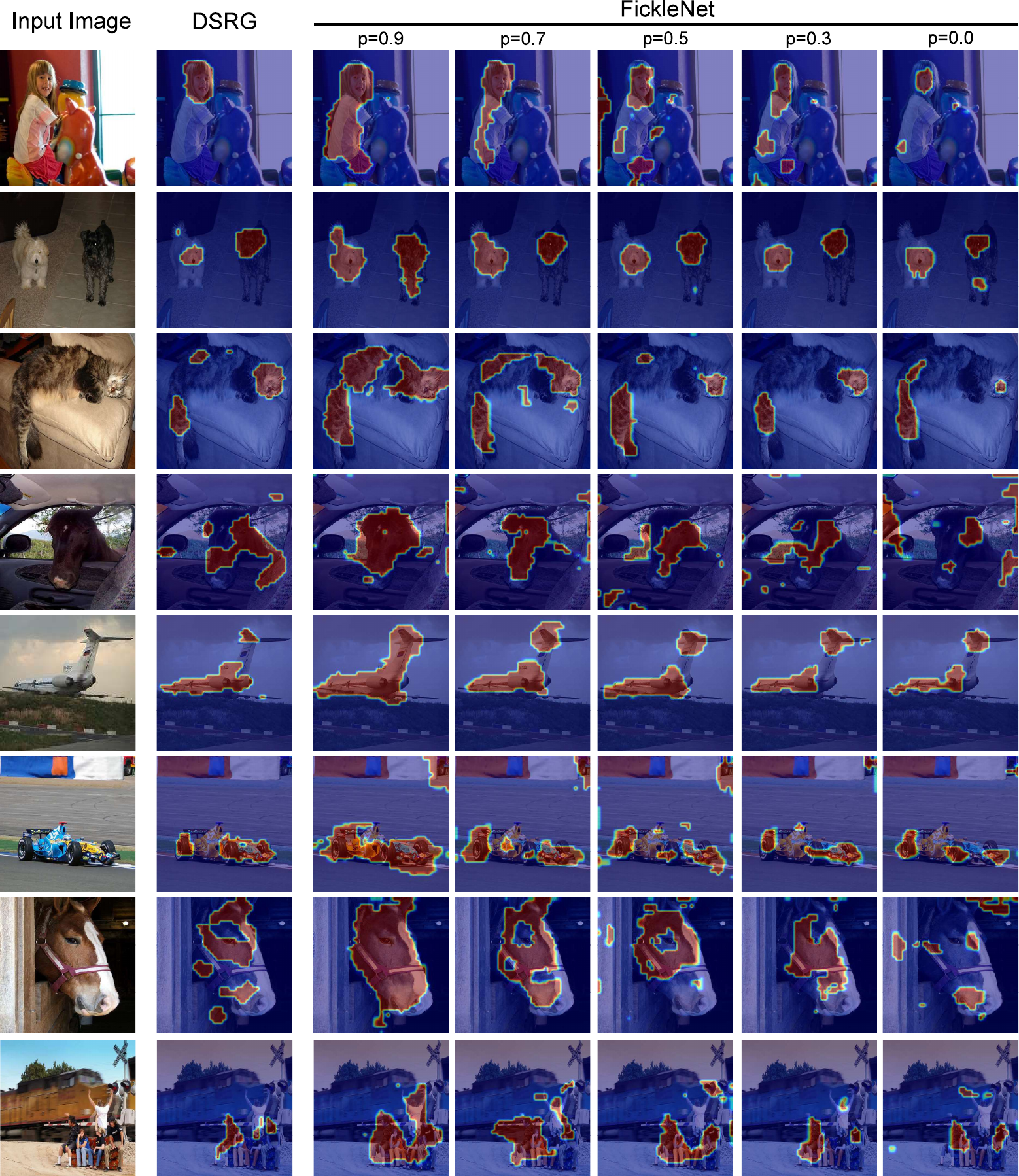}
  \caption{\label{cam_samples_supple}Localization maps from DSRG and FickleNet, with various dropout rates (p = 0 denotes a deterministic network). Localization maps of DSRG (the $2^{\text{nd}}$ column) were visualized using the publicly available DSRG localization cue.}
\end{figure*}

\begin{figure*}[t]
  \centering
  \includegraphics[width=1.0\linewidth]{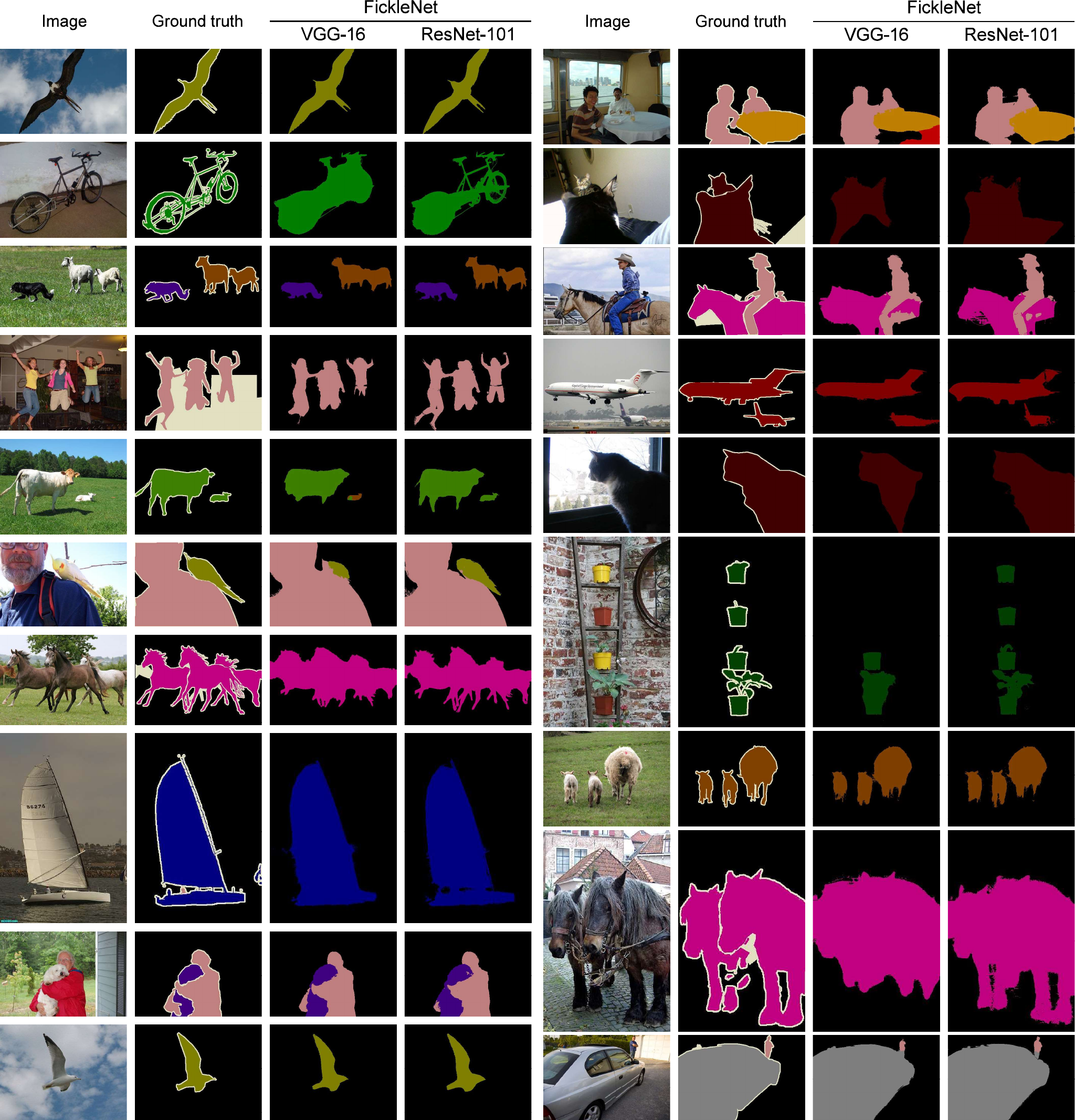}
  \caption{\label{seg_samples}Examples of predicted segmentation masks for Pascal VOC 2012 validation images of FickleNet using VGG-16 and ResNet-101.}
\end{figure*}

\noindent\textbf{Qualitative Results:} Figure~\ref{cam_samples_supple} shows additional examples of localization maps from DSRG and FickleNet for various dropout rates. Figure~\ref{seg_samples} shows the qualitative results of predicted segmentation masks in FickleNet.

\begin{table}[htbp]
\centering
  \caption{Quantitative results with and without CRF. Delta denotes the difference with and without CRF. $^{*}$ denotes the result that we reproduce.}
\resizebox{0.45\textwidth}{!}{
  
    \begin{tabular}{lccc}
    \Xhline{1pt}\\[-0.95em]

    \multicolumn{1}{c}{\multirow{2}[0]{*}{Method}} & \multicolumn{2}{c}{mIoU} & \multicolumn{1}{c}{\multirow{2}[0]{*}{Delta}} \\
    
    \multicolumn{1}{c}{} & ~with CRF~ & without CRF ~& \multicolumn{1}{c}{} \\
    \hline\hline\\[-0.95em]
    FickleNet~(ours) &  61.2     &  59.7     &  1.5\\
    DSRG~\cite{huang2018weakly}   &    \kern 0.5em 58.9$^{*}$   &  \kern 0.5em 56.5$^{*}$     &   2.4 \\
    MDC~\cite{wei2018revisiting}  &  60.4     &    57.1   &   3.3 \\
    GAIN~\cite{li2018tell}  &   55.3    &  50.8     &   4.5 \\
    \Xhline{1pt}
    \end{tabular}%
  \label{without_crf}%
  }
\end{table}%

\renewcommand{\tabcolsep}{2pt}

\begin{table*}[htbp]
  \caption{Comparison of per-class mIoU scores from DSRG, which is our backbone network, and FickleNet.}
  \centering
  \begin{adjustbox}{max width=\textwidth}
    \begin{tabular}{lccccccccccccccccccccc|c}
    
        \Xhline{1pt}

        \\[-0.95em]
    Rates~~~ & bkg& aero  & bike  & bird  & boat  & bottle & bus   & car   & cat   & chair & cow   & table & dog   & horse & motor & person & plant & sheep & sofa  & train & tv  ~ & mIOU \\
   
    \hline
    \hline
    \\[-0.9em]
   \multicolumn{22}{l}{Results on validation images:}\\
FickleNet(w/o. CRF) &   87.6    &   71.6    &   31.4    &   73.3    &   46.8    &    58.5   &    78.7   &   71.4    &   77.8    &    24.4   &   64.2    &   41.2    &    75.1   &    65.4   &   65.3    &    68.9   &   42.5    &   69.8    &   33.7    &    53.3   &    53.2  ~ &  59.7\\
    DSRG &   87.5    &   73.1    &   28.4    &   75.4    &   39.5    &    54.5  &   78.2    &   71.3    &   80.6    &   25.0    &   63.3    &    25.4   &   77.8    &    65.4   &   65.2    &   72.8    &   41.2    &    74.3   &   34.1    &   52.1    &   53.0   ~ &  59.0\\
    FickleNet &   88.1    &   75.0    &   31.3    &   75.7    &   48.8   &   60.1    &    80.0  &    72.7   &    79.6   &  25.7     &   67.3    &   42.2    &    77.1&    67.5   &   65.4    &  69.2 &  42.2   &    74.1  & 34.2  &  53.7 &   54.7~ &  61.2\\
    FickleNet~(Semi) &   90.8    &   83.9    &   34.5    &   79.9    &   53.0   &   57.4    &    85.7  &    76.7   &    81.1   &  25.1     &   69.2    &   52.1    &    77.5&    71.7   &   71.5    &  75.2 &  41.2   &    84.1  & 43.4  &  65.0 &   62.5~ &  65.8\\
    FickleNet (ResNet 101) &   89.5    &   76.6    &   32.6    &   74.6    &   51.5   &   71.1    &    83.4  &    74.4   &    83.6   &  24.1     &   73.4    &   47.4    &    78.2&    74.0   &   68.8    &  73.2 &  47.8   &    79.9  & 37.0  &  57.3 &   64.6~ &  64.9\\
    \hline
    \multicolumn{22}{l}{Results on test images:}\\
    FickleNet &   88.5    &   73.7    &   32.4    &   72.0    &   38.0   &   62.8    &    77.4  &    74.4   &    78.6   &  22.3     &   67.5    &   50.2    &    74.5&    72.1   &   77.3    &  68.8 &  52.5   &    74.8  & 41.5  &  45.5 &   55.4~ &  61.9\\
    FickleNet (ResNet 101) &   89.8    &   78.3    &   34.1    &   73.4    &   41.2   &   67.2    &    81.0  &    77.3   &    81.2   &  29.1     &   72.4    &   47.2    &    76.8&    76.5   &   76.1    &  72.9 &  56.5   &    82.9  & 43.6  &  48.7 &   64.7~ &  65.3\\
        \Xhline{1pt}

    \end{tabular}%
  \end{adjustbox}%
  \label{class-specific-results}%
\end{table*}%


\end{document}